\documentclass{article}
\usepackage{ijcai19}

\usepackage{times}
\usepackage{soul}
\usepackage{url}
\usepackage[hidelinks]{hyperref}
\usepackage[utf8]{inputenc}
\usepackage[small]{caption}
\usepackage{graphicx}
\usepackage{amsmath}
\usepackage{amssymb}
\usepackage{booktabs,tabularx}
\usepackage{algorithm}
\usepackage{algorithmic}
\usepackage{framed,multirow}

\newcommand\Tstrut{\rule{0pt}{2.6ex}}         
\newcommand\Bstrut{\rule[-0.9ex]{0pt}{0pt}}
\newcolumntype{L}[1]{>{\raggedright\let\newline\\\arraybackslash\hspace{0pt}}m{#1}}
\newcolumntype{C}[1]{>{\centering\let\newline\\\arraybackslash\hspace{0pt}}m{#1}}
\newcolumntype{R}[1]{>{\raggedleft\let\newline\\\arraybackslash\hspace{0pt}}m{#1}}

\title{CRCEN: A Generalized Cost-sensitive Neural Network Approach for Imbalanced Classification}

\author{
    Xiangrui Li \and Dongxiao Zhu
    \affiliations Department of Computer Science, Wayne State University
    \emails \{fx7219, dzhu\}@wayne.edu
}

\begin{document}

\maketitle

\begin{abstract}
Classification on imbalanced datasets is a challenging task in real-world applications. Training conventional classification algorithms directly  by minimizing classification error in this scenario can compromise model performance for minority class while optimizing performance for majority class. Traditional approaches to the imbalance problem include re-sampling and cost-sensitive methods. In this paper, we propose a neural network model with novel loss function, CRCEN, for imbalanced classification. Based on the weighted version of cross entropy loss, we provide a theoretical relation for model predicted probability, imbalance ratio and the weighting mechanism. To demonstrate the effectiveness of our proposed model, CRCEN is tested on several benchmark datasets and compared with baseline models.
\end{abstract}

\section{Introduction}
Imbalanced datasets are often encountered in real-world problems such computer vision \cite{buda2018systematic,huang2016imblcv}, healthcare informatics \cite{puru2018benchmarkmimic,li2018multinomial} and natural language processing \cite{li2011semi,liu2006speech}. In those scenarios, certain classes have abundant training data while some classes only have limited amount of data. The imbalance in training data makes learning classification models a challenging task. Conventional machine learning models treat each training sample equally and be trained by minimizing the overall classification error. Consequently, learned models bias toward correct classification in the majority class over the minority class, resulting in significant performance degradation in the minority class. This phenomenon is not desirable, particularly in domains such as predictions in intensive care units where correct classification of minority samples is critical. Hence, effective learning from imbalanced data is of great importance in machine learning applications.

To tackle the class imbalance problem, different effective strategies have been developed \cite{he2008review}. At the data level, sampling techniques are used as a pre-processing step. Training data are resampled to generate a balanced dataset, either by oversampling the minority class, or undersampling the majority class (or combine both). Then classification models are trained on the balanced data \cite{fernandez2018book}. However, sampling methods modify the original data distribution which introduces risks into the following model training: oversampling is at the risk of overfitting while undersampling loses information of majority class. To alleviate the information loss, undersampling can be further combined with ensemble methods where each base classifier is trained on undersampled balanced data.

At the algorithm level, instead of assigning equal weight on all samples, cost-sensitive learning assigns heavier costs on the misclassification of minority samples than majority ones. Classification models are then trained to minimize the total cost. In general, assignment of cost parameters heavily relies on the specific considered problems and there are no general rules for assignment. Common strategies include dynamic cost generation within boosting ensemble methods \cite{galar2012ensemblereview}, and class-wise re-weighting according to class frequency \cite{aurelio2019wce,castro2013csmlpl}. In particular, as mentioned in \cite{castro2013csmlpl}, the class-reweighting cost-sensitive methods are theoretically appealing as the class prior information can be potentially incorporated through the weighting mechanism, consequently improving learning on the minority class.

Recently, solving imbalance problem using (deep) neural networks has attracted much attention . Those works exploit deep neural network's (DNN) merit of effective feature learning to enhance predictive performances. DNNs are deployed as classifier combining with sampling techniques, cost-sensitive learning or new loss functions \cite{wang2016ijcnn,chung2015cost}. In computer vision where high-level structured information can be learned by convolutional neural networks, properties of learned feature representations are further exploited to improve DNN's performance \cite{huang2016imblcv,dong2018imblcv,khan2018cscv}.

In this paper, motivated by \cite{castro2013csmlpl}, we propose a novel cost-sensitive neural networks, Class-wise Reweighted  Cross-Entropy Network (CRCEN), to address the imbalanced binary classification problem. In CRCEN, a neural network (in our case, MLP) is used to transform the raw input features into high-level feature representations, based on which final predictions are made using sigmoid function. Different from the conventional cross entropy loss for binary classification where each training sample is equally weighted, CRCEN imposes different weights on different classes. The reweighting mechanism is capable of promoting CRCEN's learning on the minority class, enhancing the overall predictive performance. The main contributions of our proposed model are:
\begin{itemize}
	\item We propose a novel loss function of class-wise reweighted cross entropy based on neural networks to address imbalance classification problem. 
	\item We provide a theoretical derivation on the relation in sample's predicted probability (once neural network is trained), class weights of loss function and class imbalance ratio. This relation can be generalizable and hold valid for deep neural networks.
	\item We analyze the generalized relation to gain insights on model performance in imbalanced learning.
	\item We conduct extensive experiments to demonstrate the effectiveness of our method.
\end{itemize}

 The rest of the paper is organized as follows. In Section \ref{sec:relwork}, we review related works in imbalanced classification problem. Section \ref{sec:method} provide details of the proposed method and analysis on the weighting mechanism. In Section \ref{sec:exp}, our method is evaluated on several benchmark datasets. Finally, Section \ref{sec:conclu} concludes the paper with discussion.

\section{Related Work} \label{sec:relwork}
Various approaches on the class imbalance problem have been developed. Here, we focus on two widely-used approaches: sampling methods and cost-sensitive learning. This section also reviews applications of neural networks as our proposed method combines the cost-sensitive learning and MLP. For details of other methods, we refer  \cite{he2008review,he2013book,fernandez2018book} 

\paragraph{Sampling methods} Sampling techniques reduce data imbalance by either oversampling the minority class or undersampling the majority class. \cite{chawla2002smote} proposes the synthetic minority oversampling technique (SMOTE) to oversample by linearly interpolating a pair of close minority samples. ADASYN \cite{he2008adasyn} oversamples according to the local in-class density for each minority sample.  To overcome information loss of discarding samples, undersampling is often combined with ensemble methods. \cite{khoshgoftaar2007brf} explores random forest with balanced undersampled dataset in the bagging stage. \cite{liu2009ensemble} develops balanced bagging and BalanceCascade that dynamically removes easy samples from base classification models. 

\paragraph{Cost-sensitive learning} As an alternative to sampling methods, cost-sensitive (CS) learning tackle the imbalance problem by imposing heavier costs on misclassified samples. \cite{tang2009svms} develops cost-sensitive SVM with repetitive undersampling to improve the detection of informative samples. Boosting is another popular strategy, due to its internal reweighting strategy in the learning process: boosting dynamically adjusts sample weights for the next iteration according to its previous classification error \cite{freund1997boosting}. In imbalanced learning, models tends to misclassify minority samples at the early stage of boosting. With sample reweighting, minority samples are paid more attention in the later stage and hence performance on minority class can be improved \cite{sun2007csboost}. Other variants of boosting are further combined with sampling techniques to improve boosting performance \cite{galar2013eusboost,seiffert2010rusboost}. 

\paragraph{Neural networks in imbalanced learning} Neural network is a powerful machine learning model for its merit of high-level feature representation learning. \cite{dumpala2018s2s} pairs training samples and uses MLP to predict their labels simultaneously. \cite{castro2013csmlpl} develops cost-sensitive MLPs, where loss functions are squared error (L2 loss) and cross entropy respectively, both with each class's log probability reweighted by inverse class frequency.\cite{wang2016ijcnn} designs new loss functions that encourage equal classification error for majority and minority class for image classification. Additionally, with effective feature extraction of convolutional neural networks, \cite{huang2016imblcv,khan2018cscv,dong2018imblcv} solve the imbalanced problem in computer vision by exploiting the structured information in high-level features.

Our proposed CRCEN for binary imbalanced classification is a neural network based CS learning method motivated by CS's theoretical appeal, as mentioned in \cite{castro2013csmlpl}. The loss function in  CRCEN is the class-wise reweighted cross entropy (CE), and the choice of weight is related to the model's predictive behavior.

Previous work that is closest to our work in this paper  is \cite{castro2013csmlpl}. \cite{castro2013csmlpl} proposes cost-sensitive MLP (CSMLP) and its loss function is the class-wise weighted squared error loss (i.e. treat classification problem as regression, and label is coded as $+1$ for majority class and $-1$ for minority class). Under that specific setting, they derive the same weighting strategy of inverse class frequency to improve learning on the minority class. However, dependence of their weight derivation on the label coding is unclear; it is also difficult to understand the effect if a different weight setting is used. Different from their regression treatment, CRCEN takes the natural probabilistic approach (CE loss) to imbalanced classification problem, which does not rely on a specific label coding. Under some moderate conditions, a relation on MLP's predicted probability, weight choice and class imbalance ratio is theoretically derived. That relation is further qualitatively analyzed to gain insights on MLP's predictive performance as well as understand the effect of class weights from a probabilistic perspective. We have noticed that \cite{aurelio2019wce} recently uses weighted CE loss for imbalanced learning, incorporating prior class information heuristically. This is a special case of CRCEN. Furthermore, we provide a theoretical support for this particular weight choice under CRCEN framework.

\section{Proposed Method} \label{sec:method}
\paragraph{Notation} Let $S = \{(\boldsymbol{x}_i, y_i): i=1,\cdots,N\}$ be the imbalanced set of $N$ training samples,  where $\boldsymbol{x}_i\in \textbf{R}^p$ is the $p$-dimensional feature vector and $y_i\in \{0,1\}$ is the class label.We use $y=1$ to represent the minority class and $y=0$ the majority. Further, $S_0 = \{(x_j, y_j):y_j=0, j=1,\cdots,N_0 \}$ and $S_1 = \{(x_i,y_i): y_i=1,i=\cdots, N_1\}$ represent the sample sets of majority and minority class respectively with $N_0 \gg N_1$ and $N=N_0+N_1$. Imbalance ratio (IR) is defined as $N_0/N_1 \gg 1$.  

\subsection{The CRCEN Loss Function}
For the binary classification problem, minimizing cross entropy (CE) (equivalent to maximum likelihood) is an attractive approach due to its modeling of uncertainty, where each training sample is weighted equally. In imbalanced learning, direct use of CE often leads to poor predictive performance. The learning algorithm favors correct classification of samples in the majority class and tends to misclassify the minority samples. 

To overcome this limitation and improve correct classification on the minority class, we propose CRCEN to optimize the following class-wise reweighted version of CE loss:
\begin{equation} \label{eq:CRCEN}
\begin{split}
L(\boldsymbol{\theta}) = & -\sum_{i\in S} \lambda y_i \log p_i(\boldsymbol{\theta}) +(1-\lambda)(1-y_i) \log (1-p_i(\boldsymbol{\theta}))\\
 =& -\lambda \sum_{i\in S_1} \log p_i (\boldsymbol{\theta}) - (1-\lambda)\sum_{j\in S_0}\log(1-p_j(\boldsymbol{\theta}))\\
  = & \lambda L_1(\boldsymbol{\theta}) + (1-\lambda)L_0(\boldsymbol{\theta}),
\end{split}
\end{equation} 
or the regularized version of $L(\boldsymbol{\theta})$ (such as $L_2$ norm of $\boldsymbol{\theta}$):
$$ L(\boldsymbol{\theta}) + \beta\Omega(\boldsymbol{\theta}),$$
where $p_i(\boldsymbol{\theta})$ is the probability $p(y=1|\boldsymbol{x}_i)$ of $\boldsymbol{x}_i$ belonging to the minority class, modeled by a neural network $f_{\boldsymbol{\theta}}(\boldsymbol{x}_i)$ with parameter vector $\boldsymbol{\theta}$, $0<\lambda<1$ is the weight parameter, $\beta$ the tuning parameter of $\boldsymbol{\Omega}$. In this paper, $f_{\boldsymbol{\theta}}$ is chosen as MLP.

Note that when $\lambda=1/2$, CRCEN reduces to the conventional CE loss. Then in the optimization of $L(\boldsymbol{\theta})$ using gradient descent, the gradient is dominated by the error signal of majority class and decision boundary will be pushed towards minority class. As a consequence, $f_{\boldsymbol{\theta}}$ is likely to have high classification error for minority class. Hence, when greater than $1/2$, $\lambda$ can be viewed intuitively to strengthen the error signal of minority class in the gradient. Alternatively, optimizing Equation \ref{eq:CRCEN} is equivalent to optimizing conventional CE loss, but where training data are rebalanced by simply duplicating each minority sample $\lambda/(1-\lambda)$ times.

\subsection{A Key Equation for Weight $\lambda$}
Reweighting minority samples has been effectively applied in practices. A common strategy is to upweight minority samples by the imbalance ratio IR. In this section, we investigate the theoretical aspect of this weighting mechanism, \textit{with neural network being the predictive models}. As we shall see, the weight $\lambda$ connects the imbalance ratio with (expected) sample's predicted probability $p(y=1|\boldsymbol{x}_{\text{min}}) = f_{\boldsymbol{\theta}}(\boldsymbol{x}_{\text{min}})$ and $p(y=0|\boldsymbol{x}_{\text{maj}}) = 1 - f_{\boldsymbol{\theta}}(\boldsymbol{x}_{\text{maj}})$ , where $\boldsymbol{x}_{\text{min}}$ and $\boldsymbol{x}_{\text{maj}}$ are minority and majority samples respectively. In the following subsections, we remove $\boldsymbol{\theta}$ in analysis for notational brevity.

Assume the output layer of MLP $f$ consists of only one neuron, then the predicted probability is 
\begin{equation} \label{eq:predprob}
\begin{split}
 p(y=1|\boldsymbol{x}) &= f(\boldsymbol{x}) = \frac{1}{1+\exp(-o_{\boldsymbol{x}})},\\
 o_{\boldsymbol{x}} &= b+\boldsymbol{W}^T \cdot \boldsymbol{h}_{\boldsymbol{x}},
\end{split}
\end{equation}
where $o$ is the input to the output neuron, $\boldsymbol{h_{\boldsymbol{x}}}$ is the feature embedding of $\boldsymbol{x}$ from the last hidden layer, $\boldsymbol{W}$ is a parameter subvector of $\boldsymbol{\theta}$ for the output neuron and $b$ be the corresponding bias term.

After MLP is trained, the loss function $L(\boldsymbol{\theta})$ is minimized with optimal solution $\boldsymbol{\theta}^*$. Here, we focus on $L(\boldsymbol{\theta})$ for simplicity. By optimization theory, we have a necessary condition at $\boldsymbol{\theta} = \boldsymbol{\theta}^*$:
\begin{equation*}
\begin{split}
\frac{\partial L}{\partial \boldsymbol{\theta}}  =\boldsymbol{0}
\iff   \lambda \frac{\partial L_1}{\partial \boldsymbol{\theta}} + (1-\lambda)\frac{\partial L_0}{\partial \boldsymbol{\theta}}= \boldsymbol{0}.
\end{split}
\end{equation*}

For simplicity, we consider one component $w$ of $\boldsymbol{\theta}$, at $w=w^*$:
\begin{align} \label{eq:necessaryCon}
&\lambda \frac{\partial L_1}{\partial w} + (1-\lambda)\frac{\partial L_0}{\partial w} = 0,\\
& \frac{\partial L_1}{\partial w} = - \sum_{i\in S_1}\frac{1}{p_i}\frac{\partial p_i}{\partial w}, \nonumber \\
& \frac{\partial L_0}{\partial w} = \sum_{j \in S_0}\frac{1}{1-p_j}\frac{\partial p_j}{\partial w}, \nonumber
\end{align}
where $p_i = f(\boldsymbol{x}_i) = p(y=1|\boldsymbol{x}_i)$ is the predicted probability of sample $\boldsymbol{x}_i$ belonging to minority class.

Since for sigmoid function $\sigma(x) = 1/(1+\exp(-x))$, its derivative is 
$$ \sigma'(x) = \sigma(x)(1-\sigma(x)).$$
Using chain rule, we have the following:
\begin{equation} \label{eq:chain}
\begin{split}
&\frac{\partial p_i}{\partial w} = \frac{\partial p_i}{\partial o_i}\frac{\partial o_i}{\partial w} = p_i(1-p_i)\frac{\partial o_i}{\partial w},\\
&\frac{\partial p_j}{\partial w} = \frac{\partial p_j}{\partial o_j}\frac{\partial o_j}{\partial w} = p_j(1-p_j)\frac{\partial o_j}{\partial w}.
\end{split}
\end{equation}

Plug Equation (\ref{eq:chain}) into Equation (\ref{eq:necessaryCon}), we have:
\begin{equation} \label{eq:compNecCond}
-\lambda \sum_{i\in S_1} (1-p_i)\frac{\partial o_i}{\partial w} + (1-\lambda)\sum_{j\in S_0} p_j \frac{\partial o_j}{\partial w} = 0.
\end{equation}

Since Equation (\ref{eq:compNecCond}) holds for any component of parameter $\boldsymbol{\theta}$, we specifically consider the case $w=b$ for the bias term $b$ in Equation (\ref{eq:predprob}). Hence $\partial o_i / \partial b = \partial o_j / \partial b = 1$. From Equation (\ref{eq:compNecCond}), we obtain the key equation of CRCEN:

\begin{equation} \label{eq:KeyEq1}
\frac{\sum_{i\in S_1} 1-p_i}{\sum_{j \in S_0} p_j }= \frac{1-\lambda}{\lambda},
\end{equation} 
which reveals the relation of weight $\lambda$, training sample's predicted probability $p_i (i\in S_1)$ and $p_j (j\in S_0)$ for the minority and majority class, after the neural network is trained.

In the neural network training, $L_2$ regularization is often applied to prevent overfitting. If the bias term $b$ is not regularized in $L_2$ term (which is the case usually), Equation (\ref{eq:KeyEq1}) still holds. Let $\boldsymbol{\theta} = (\boldsymbol{W}_h, \boldsymbol{b})$, where $\boldsymbol{W}_h$ is the parameter vector for hidden layers and $\boldsymbol{b}$ the vector of all bias terms. $L_2$ regularization $\Omega(\boldsymbol{\theta})= ||\boldsymbol{W}_h||_2^2$. Then we have for any component $b$ of $\boldsymbol{b}$,
$\partial \Omega/\partial b =0$. Hence Equation (\ref{eq:compNecCond}) holds for $w=b$.

\subsection{Theoretical Analysis on Equation (\ref{eq:KeyEq1})}
The relation given by Equation (\ref{eq:KeyEq1}) depends on the training dataset $S=S_1 \cup S_0$. For theoretical analysis, we make a moderate assumption that, in both the training and testing data, the majority and minority samples are generated from the same class-conditional distribution $ P_1(\boldsymbol{x}|y=1)$ and $ P_1(\boldsymbol{x}|y=0)$ respectively (i.e no distribution shift between training and testing). Hence, Equation (\ref{eq:KeyEq1}) can be generalized as
\begin{align} \label{eq:KeyEq2}
& \frac{1-\bar{p}_1}{\bar{p}_0} \approx \frac{N_0(1-\lambda)}{N_1\lambda},\\
& \sum_{j\in S_0} p_j \approx N_0 \bar{p}_0, \sum_{i\in S_1} 1-p_i \approx N_1 (1-\bar{p}_1), \nonumber \\
& \bar{p}_1 = \text{\textbf{E}}_{\boldsymbol{x}\sim P_1(\boldsymbol{x}|y=1)} f_{\boldsymbol{\theta}^*}(\boldsymbol{x}), \nonumber \\
& \bar{p}_0 = \text{\textbf{E}}_{\boldsymbol{x}\sim P_0(\boldsymbol{x}|y=0)} f_{\boldsymbol{\theta}^*}(\boldsymbol{x}), \nonumber
\end{align}
where  $P(\boldsymbol{x}|y=1)$ and $P(\boldsymbol{x}|y=0)$ are the distributions of the minority and majority class, \textbf{E} is the expectation operator. Hence, $1- \bar{p}_1$ is the expected probability of a minority sample with which the trained neural network predicts it in the majority class; $\bar{p}_0$ be the expected probability of a majority sample being predicted as a minority sample. Here, we use $f_{\boldsymbol{\theta}^*}$ to emphasize the dependence of (\ref{eq:KeyEq2}) on the trained neural network model. Note that Equation (\ref{eq:KeyEq2}) is a general relation regardless of data imbalance.

Predictive performance of the classifier involves a decision making process given $p_i$. The conventional approach is to set a probability threshold $t$, such that $y=1$ if $f_{\boldsymbol{\theta}^*}(x)> t$, $y=0$ otherwise. Here, we take $t=0.5$ for the following analysis to understand model performance. By assuming that training and testing data follow the same distribution, Equation (\ref{eq:KeyEq2}) is generalizable from training to testing.

\subsubsection{When $\boldsymbol{\lambda=1/2}$}
CRCEN reduces to the conventional cross entropy loss. When imbalance ratio is high in the training data, say $N_0/N_1 = 10$, Equation (\ref{eq:KeyEq2}) is $1-\bar{p}_1 = 10 \bar{p}_0.$
Since $\bar{p}_1 \geq 0$, we must have
$$\bar{p}_0 \leq 0.1 \iff 1-\bar{p}_0 \geq 0.9.$$
If $t=0.5$ is the decision making threshold, this implies that the trained neural network can correctly predict a majority sample, confidently (at least) with probability 0.9, on average. 

For prediction on minority class, model performance is more complex. We illustrate the idea with two cases for $\bar{p}_0$. Since $\bar{p}_0 \leq 0.1$, again, take $t=0.5$,
\begin{itemize}
	\item if $\bar{p}_0 = 0.08$, then we have $\bar{p}_1=0.2$. That implies the predicted probability of a minority sample being minority is 0.2 on average. Hence, the classifier must misclassify most minority samples ($0.2 < 0.5$), resulting in very poor predictive accuracy for minority class.
	\item if $\bar{p}_0 = 0.04$, then $\bar{p}_1 = 0.6$. Hence, the classifier can achieve good performance for minority samples. But the extent of ``goodness" depends on the trained network $f_{\boldsymbol{\theta}^*}$ and $P_0(\boldsymbol{x}|y=1)$, i.e. the variance $\text{V}_{\boldsymbol{x}\sim P_1(\boldsymbol{x}|y=1)}(f_{\boldsymbol{\theta}^*}(\boldsymbol{x}))$. If the variance is small (relative to the average distance from the decision boundary in the latent feature space), classifier can still achieve very high accuracy. If large, performance would degrade.
\end{itemize}

Geometrically, the imbalance of training data would push decision boundary toward to minority class in the latent feature space learned by neural network.

The analysis above theoretically explains why classifier always has good performance for majority class, as well as how performance for minority class is connected with model training and data distribution in imbalanced learning.

\subsubsection{When $\boldsymbol{\lambda=N_0/(N_0+N_1)}$} This strategy is equivalent to the empirical weighting by inverse class frequency deployed in \cite{aurelio2019wce,castro2013csmlpl}. With this choice of $\lambda$, Equation (\ref{eq:KeyEq2}) is $\bar{p}_1 = 1-\bar{p}_0$. That is, the probability of predicting a minority sample as minority is equal to the probability of predicting a majority sample as majority. Taking  $t=0.5$, good predictive performance for both minority and majority class is guaranteed. However, the extent of ``goodness" depends on $\text{V}_{\boldsymbol{x}\sim P_1(\boldsymbol{x}|y=1)}(f_{\boldsymbol{\theta}^*}(\boldsymbol{x}))$ and $\text{V}_{\boldsymbol{x}\sim P_0(\boldsymbol{x}|y=0)}(f_{\boldsymbol{\theta}^*}(\boldsymbol{x}))$.

\subsubsection{For General Choice of $\boldsymbol{\lambda}$}
$\lambda$ is the parameter controlling the ratio between probabilities $1-\bar{p}_1$ and $\bar{p}_0$. For example, when $\lambda = 2N_0/(N_1+2N_0)$, $1-\bar{p}_1 = \bar{p}_0/2$. The loss function in CRCEN then deploys a  weight setting equivalent to:
$$ 2N_0L_1(\boldsymbol{\theta}) + N_1L_0(\boldsymbol{\theta}).$$ 
Since $0 \leq \bar{p}_0 \leq 1$, $\bar{p}_1\geq 0.5$. We are guaranteed to have good predictive performance for minority class (when $t=0.5$). Assume $\bar{p}_0 =0.4$ \footnote{The true value of $\bar{p}_0$ is generally unknown as it depends on the true data distribution. } (i.e. $1-\bar{p}_0=0.6)$, we obtain $\bar{p}_1=0.8$, which implies the prediction accuracy for minority class can be possibly boosted at the cost of a small accuracy degradation, but still maintaining good performance, for majority class. For the exact relation, we plan to investigate this problem in our future studies.

\begin{table}[t]
	\centering
	\begin{tabular}{cccc}  
		\toprule
		$\lambda$  & $\frac{1}{2}$  & $\frac{N_0}{N_0+N_1}$  & $\frac{2N_0}{(2N_0+N_1)}$ \\
		\midrule
		RHS & 10 & 1 & 0.5  \\
		\hline
		\multirow{2}{*}{LHS (Sim1)} & 10.05 \Tstrut  & 1.00 \Tstrut  & 0.50 \Tstrut \\
		& (1.13) & (0.09) & (0.04)\\
		
		\multirow{2}{*}{LHS (Sim2)} & 10.12 \Tstrut  & 1.01 \Tstrut  & 0.50 \Tstrut \\
		& (0.67) & (0.05) & (0.03)\\
		\bottomrule
	\end{tabular}
	\caption{Simulation results (along with standard deviation) for Equation (\ref{eq:KeyEq2}) over 100 runs. RHS represents theoretical value on the right-hand side of (\ref{eq:KeyEq2}); LHS the simulated value on the left hand side.}
	\label{tab:sim}
\end{table}

\subsection{Simulations for Correctness of Equation (\ref{eq:KeyEq2})}
In order to check the correctness of Equation (\ref{eq:KeyEq2}), we conduct simulations under two settings. The imbalance ratio is 10 in training data ($N_1=1000, N_0=10000$), testing data size is $(1000, 1000)$; both training and testing data follow the same data distribution.
\begin{itemize}
	\item Sim1: $P_1(x|y=1) =\mathcal{N}(-1.5, 1)+\mathcal{U}(0,0.5)$, $P_2(x|y=0) = \mathcal{N}(1.5, 1) + \mathcal{U}(-0.5, 0)$. Logistic regression is fitted (as a special case of MLP).
	\item Sim2: $P_1(\boldsymbol{x}|y=1) = \mathcal{N}(\boldsymbol{\mu}_1, \boldsymbol{\sigma}_1)$, $P_0(\boldsymbol{x}|y=0) = \mathcal{N}(\boldsymbol{\mu}_0, \boldsymbol{\sigma}_0)$, where $\boldsymbol{\mu}_1 = (0,0,0)$, $\boldsymbol{\mu}_0 = (1,1,1)$, $\boldsymbol{\sigma}_1 = 1.2\boldsymbol{I}$, $\boldsymbol{\sigma}_0 = \boldsymbol{I}$. A one-hidden-layer MLP of layer size $(3,10,1)$ and sigmoid activation is fitted. 
\end{itemize}

$\lambda$ is the weight used in CRCEN and the latter is then trained on the training data. The predicted probability for testing data is calculated and used in the LHS of Equation (\ref{eq:KeyEq2}) to approximate $\bar{p}_1$ and $\bar{p}_0$. Table \ref{tab:sim} shows simulation results under three $\lambda$ settings. We see from the table that simulated values match with the theoretical values accurately, demonstrating the correctness of Equation (\ref{eq:KeyEq2}).

\begin{table}
	\centering
	\begin{tabular}{cccc}  
		\toprule
		Dataset  & \# Sample & \# Feature & IR \\
		\midrule
		Abalone       & 4177 & 10 & 9.7    \\
		Coil & 9822 & 85 & 16 \\
		Satimage & 6435 & 36 & 9.3 \\
		Scene & 2407 & 294 & 13 \\
		Solar & 1389& 32 &19 \\
		UScrime & 1994 & 100 & 12\\
		\bottomrule
	\end{tabular}
	\caption{Details of datasets. IR represents imbalance ratio between majority class and minority class.}
	\label{tab:dataset}
\end{table}

\begin{table*}[th]
	\centering
	\begin{tabular}{|c|c|c|c|c|c|c|c|c|c|c|c|c|}
		\hline
		& \multicolumn{2}{|c|}{Abalone} \Tstrut & \multicolumn{2}{|c|}{Coil} \Tstrut & \multicolumn{2}{|c|}{Satimage} \Tstrut & \multicolumn{2}{|c|}{Scene} \Tstrut & \multicolumn{2}{|c|}{Solar} \Tstrut & \multicolumn{2}{|c|}{UScrime}\Bstrut \\ 
		\hline
		& F1 \Tstrut & Gm \Tstrut & F1 \Tstrut & Gm \Tstrut & F1 \Tstrut & Gm \Tstrut & F1 \Tstrut & Gm \Tstrut  & F1 \Tstrut & Gm \Tstrut & F1 \Tstrut & Gm \Tstrut \\
		\hline
		MLP \Tstrut & 0 & 0 & 0.107 & 0.273 & 0.554 & 0.668 & 0.211 & 0.341 &0 & 0 & 0.475 & 0.597 \\
		\hline
		CSMLP \Tstrut & 0.392  & 0.798  & 0.200 & 0.658 & 0.611 & 0.890 & 0.259 & 0.662 & 0.240 & 0.655 & 0.472 & 0.707 \\
		\hline
		CRCEN \Tstrut & \textbf{0.400}  & \textbf{0.808} & \textbf{0.205} & \textbf{0.678} & \textbf{0.626} & 0.894 & \textbf{0.298} & 0.701 & \textbf{0.264} & 0.690 & \textbf{0.525} & 0.726 \\
		\hline 
		ADASYN \Tstrut & 0.389  & 0.804 & 0.166 & 0.411 & 0.583 & \textbf{0.898} & 0.253 & 0.512 & 0.240 & 0.643 & 0.448 & 0.652 \\
		\hline  
		SMOTE \Tstrut & \textbf{0.400}  & \textbf{0.808} & 0.161 & 0.406 & 0.624 & 0.891 & 0.234 & 0.487 & 0.232 & 0.647 & 0.466 & 0.675\\
		\hline
		RUSB \Tstrut & 0.385   & 0.792 & 0.163 & 0.624 & 0.518 & 0.874 & 0.163 & 0.561 & 0.174 & \textbf{0.705} & 0.407 & 0.831 \\
		\hline
		EE \Tstrut & 0.377  & 0.789  & 0.200 & 0.670 & 0.537 & 0.869 & 0.256 & 0.707 & 0.169 & 0.670 & 0.434 & 0.858\\
		\hline
		BRF \Tstrut & 0.393   & 0.794 & 0.196 & 0.674 & 0.579 & 0.890 & 0.256 & \textbf{0.709} & 0.180 & 0.699 & 0.459 & \textbf{0.865} \\
		\hline	
	\end{tabular}
	\caption{Predictive performance on the testing data. F1 is F-measure and Gm is G-mean.}
	\label{tab:performance}
\end{table*}

\section{Experiments} \label{sec:exp}
In this section, we evaluate CRCEN on real-world datasets with various imbalance degrees that are widely used in imbalanced learning. All datasets tested in the experiments are extracted from the ``imbalanced-learn" Python package \cite{lemaitre2017imbalanced}. Table \ref{tab:dataset} shows the details of the datasets used in the experiments.

For performance comparison, we test several baseline models for imbalanced learning, including sampling methods SMOTE \cite{chawla2002smote} and ADASYN \cite{he2008adasyn}, ensemble methods Rusboost (RUSB) \cite{seiffert2010rusboost}, balanced random forest (BRF) \cite{khoshgoftaar2007brf} and EasyEnsemble (EE) \cite{liu2009ensemble}, and neural network-based cost-sensitive method CSMLP \cite{castro2013csmlpl}. As sampling methods are a data preprocessing step, we use MLP as the classifier after training data are re-sampled. For ensemble methods, default base classifiers are used. In addition, a MLP classifier trained on the original data with conventional cross entropy loss is tested. Most methods are implemented in imbalanced-learn and sklearn package. For our proposed CRCEN, we use MLP as the classifer and implement it in Pytorch. All MLPs are fixed with 3 layers with number of hidden neurons selected for each dataset.

In the experiments, each dataset is divided into training and testing sets by a stratified split 0.75/0.25 to ensure imbalance ratio is maintained. All models are trained on the training data and model performance is evaluated on the testing data. We repeats train/test split 4 times. We select model parameters (including $L_2$ regularization, number of hidden neurons) using 4-fold cross validation with a grid search on the training data in the first run of the experiment and then fix them in the subsequent runs. This procedure is used to test model robustness to the variations in train/test splits.

As the overall accuracy is known misleading for imbalanced datasets, we use F-measure (F1) and Gm as evaluation metrics. F-measure and G-mean(Gm) are defined as follows:
\begin{equation*}
\begin{split}
\text{F1} &= \frac{2\cdot \text{Precision}\cdot\text{Recall}}{\text{Precision}+\text{Recall}},\\
\text{Gm} &= \sqrt{\text{Recall}\cdot\text{Specificity}},
\end{split}
\end{equation*}
where Precision =TP/(TP+FP), Recall = TP/(TP+FN) and Specificity = TN/(FP+TN),
TP represents true positive, FP false positive, FN false negative and TN ture negative. 

\subsection{Predictive Results}
Predictive performance on the testing data is reported in Table \ref{tab:performance}. As we can see from the table, CRCEN has an overall better performance than other methods. Compared with MLP trained directly on the training data, all methods achieve great improvements in G-mean. Since MLP tends to misclassify minority samples into majority class, this results in a large error and low recall score for minority class. Consequently, the G-mean of MLP is low. On the contrary, this demonstrates that all those techniques can effectively strengthen classifier's detection for minority class and improve overall performance. CRCEN and CSMLP are both cost-sensitive methods based on MLP with the same weighting mechanism. We see that in the experiments CRCEN has slightly better but comparable performances with CSMLP. As explained in previous sections, both methods increase learning of minority samples with theoretical guarantees, by learning a balanced boundary between two classes. However, CRCEN's probabilistic approach is more appropriate and effective for classification problems \cite{lee2015deeply}. Rusboost (RUSB) is an ensemble cost-sensitive method that costs are dynamically assigned to misclassified samples. From the table, RUSB's performance is not as good as CRCEN and CSMLP. By checking its prediction (results not shown here), RUSB achieves highest classification accuracy for minority class, at the  cost of significant amount of misclassification in majority class. This is because samples on the boundary can be easily misclassified, RUSB would assign a large cost to the boundary samples. Consequently, decision boundary will be pushed towards majority class and results in decreasing in precision and specificity, hence degradation in overall performance (F1 and G-mean).

\subsection{Effect of $\boldsymbol{\lambda}$}
In many real-world applications, correct classification of minority samples ($y=1$) is of high priority over misclassification of majority samples ($y=0$), making imbalanced learning extremely useful. In terms of evaluation metrics, a predictive model that has high recall is preferred. For cost-sensitive learning, imposing more cost on minority samples would generally improve recall. However, since there is tradeoff between recall and precision (or specificity), higher cost will decrease precision and (or specificity). Under CRCEN, the weight parameter $\lambda$ controls the tradeoff. In this section, we investigate this relation.

 \begin{figure}[!h]
	\includegraphics[scale = 0.52]{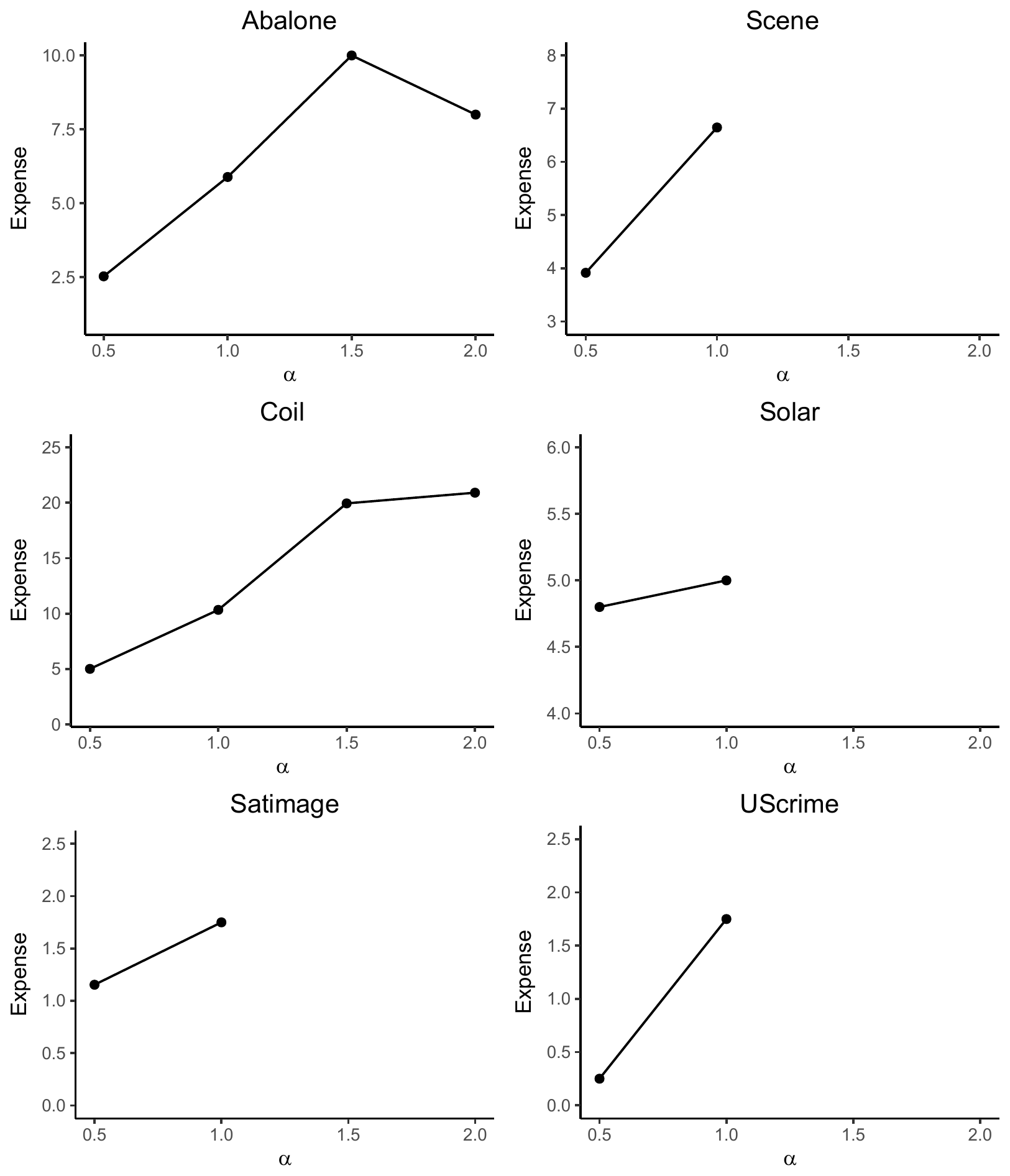}
	\caption{Expense plot for different $\lambda$ settings corresponding to $\alpha$s. Expense measures the tradeoff between FNs and FPs when heavier cost is added to minority class in CRCEN. When $\alpha=0.5$, plot corresponds to the expense from conventional cross entropy loss ($\lambda=1/2$) to cost sensitive loss CRCEN.}
	\label{fig:expplot}
\end{figure}

We set $\lambda =\frac{1}{2}$ and $ \frac{\alpha N_0}{(\alpha N_0 + N_1)}$, $\alpha=(0.5,1,1.5,2)$ to see to what extent of increase in weight can lead to performance gain on classifying the minority class, and how much performance loss on magnitude of the majority class. Note that when $\lambda=1/2$, CRCEN is equivalent to conventional cross entropy loss. For each value of $\alpha$, we train a MLP for classification. Once the models are trained, predictions are made on testing data. To quantify the tradeoff between recall and precision, we define \textit{\textbf{expense}} as the ratio of change of FPs to that of FNs between two consecutive $\alpha$s (since $\alpha$ increases, number of FN decreases and FP increases). For example, $(32,158)$ and $(13,263)$ are two pairs of (FN, FP) corresponding to $\alpha=0.5$ and $\alpha=1$ respectively, predicted on the testing data of Abalone. Then the budget is $(263-158)/(32-13) = 5.53$.  This ratio can be viewed as the budget that one can afford for detecting a false negative sample. The decision making threshold is $t=0.5$

Figure \ref{fig:expplot} shows the expense plots for six datasets. From the table, we can observe a trend that the expense increases as $\alpha$ (namely $\lambda$) increases. This implies that by imposing heavier costs, we can improve detection of true positives however at an increasing cost of false positives. When $\alpha>1$, in Scene, Solar, Satimage and UScrime datasets, heavier costs don't improve classifier's performance on minority class. With limited amount of training data, this in turn increases the risk of overfitting, resulting in performance degradation. We see that for datasets Satimage and UScrime, the expense is relatively small (less than 2 when $\alpha=1$), compared with Abalone, Scene, Coil and Solar. In Table \ref{tab:performance}, MLP already has good performance in those two datasets. With a low expense and high class imbalance ratio, cost-sensitive can further improve the model overall performance, as confirmed in Table \ref{tab:performance}. From the same perspective, CRCEN is also effective for more complex datasets Abalone, scene, Coil and Solar.

\section{Conclusion and Discussion} \label{sec:conclu}
In this paper, we proposed a novel neural-network based model CRCEN for imbalanced learning problem. The object function in CRCEN is a class-wise reweighed version of the cross entropy loss. With this simple form, under some mild conditions, we derive a non-trivial probabilistic relation that can help us understand model's predictive behavior. When the weights are set to inverse class frequency as a heuristic, the derived relation provides explanation for the effectiveness of this approach with theoretical guarantees. Extensive experiments are conducted to demonstrate the effectiveness of CRCEN. For future studies, we plan to investigate the relation for a general choice of $\lambda$ to understand how model performance is affected accordingly.

\small
\bibliographystyle{named}
\bibliography{CRCENref}

\end{document}